# Optimizing Dialogue Management with Reinforcement Learning: Experiments with the NJFun System

**Satinder Singh**                                            BAVEJA@CS.COLORADO.EDU
*Syntek Capital*
*New York, NY 10019*

**Diane Litman**                                              LITMAN@CS.PITT.EDU
*Department of Computer Science and LRDC*
*University of Pittsburgh*
*Pittsburgh, PA 15260*

**Michael Kearns**                                           MKEARNS@CIS.UPENN.EDU
*Department of Computer and Information Science*
*University of Pennsylvania*
*Philadelphia, PA 19104*

**Marilyn Walker**                                           WALKER@RESEARCH.ATT.COM
*AT&T Labs - Research*
*Florham Park, NJ 07932*

## Abstract

Designing the dialogue policy of a spoken dialogue system involves many nontrivial choices. This paper presents a reinforcement learning approach for automatically optimizing a dialogue policy, which addresses the technical challenges in applying reinforcement learning to a working dialogue system with human users. We report on the design, construction and empirical evaluation of NJFun, an experimental spoken dialogue system that provides users with access to information about fun things to do in New Jersey. Our results show that by optimizing its performance via reinforcement learning, NJFun measurably improves system performance.

## 1. Introduction

Recent advances in spoken language understanding have made it possible to develop dialogue systems for many applications. The role of the dialogue manager in such systems is to interact in a natural way to help the user complete the tasks that the system is designed to support. Typically, an expert designs a dialogue management policy by hand, and has to make many nontrivial design choices. It is often difficult to assess the ramifications of these choices because the performance of a dialogue policy depends on many other factors, such as the user population, the robustness of the automatic speech recognizer (ASR), and task difficulty (Kamm, Litman, & Walker, 1998; Danieli & Gerbino, 1995). This paper applies reinforcement learning (RL) to *automatically* learn design choices that optimize system performance for a chosen performance measure (Levin, Pieraccini, & Eckert, 2000; Walker, Fromer, & Narayanan, 1998a).

S1: Welcome to NJFun. How may I help you?
U1: I'd like to find um winetasting in Lambertville in the morning.
    (*ASR: I'd like to find out wineries the in the Lambertville in the morning.*)
S2: Did you say you are interested in Lambertville?
U2: Yes.
S3: Did you say you want to go in the morning?
U3: Yes.
S4: I found a winery near Lambertville that is open in the morning.
    It is the Poor Richard's Winery in Lambertville.
    Thank you for using the system.
    Please give me feedback by saying 'good', 'so-so', or 'bad'.
U4: Good.
S5: Goodbye and have a nice day!

Figure 1: An example dialogue with NJFun.

Consider the spoken dialogue system named NJFun, which we implemented to provide telephone access to a database of activities in New Jersey. A sample dialogue with NJFun is shown in Figure 1, with system utterances labeled S$i$ and user utterances labeled U$i$. In this dialogue, by starting with the open-ended greeting "How may I help you?", the system lets the *user* take the *initiative* in providing information about the activity they are interested in. User responses in such cases may be relatively unconstrained. In contrast, the *system* could take the initiative by saying the more restrictive phrase "Please tell me the location you are interested in", thus constraining the user to provide information about the location of the activity. Which of these contrasting choices of user or system initiative is superior may depend strongly on the properties of the underlying and imperfect ASR, the population of users, as well as the dialogue so far. This choice of initiative occurs repeatedly throughout a dialogue, and is but one example of a class of difficult design decisions.

In the main, previous research has treated the specification of the dialogue management policy as an iterative design problem: several versions of a system are created (where each version uses a single dialogue policy, intuitively designed by an expert), dialogue corpora are collected with human users interacting with different versions of the system, a number of evaluation metrics are collected for each dialogue, and the different versions are statistically compared (Danieli & Gerbino, 1995; Sanderman, Sturm, den Os, Boves, & Cremers, 1998; Kamm, 1995; Walker, Litman, Kamm, & Abella, 1998b). Due to the costs of experimentation, only a handful of policies are usually explored in any one experiment. Yet, many thousands of reasonable dialogue policies are typically possible. In NJFun, for example, there is a search space of $2^{42}$ potential dialogue polices, as will be detailed below.

Recent work has suggested that a dialogue policy can be designed using the formalisms of Markov decision processes (MDPs) and reinforcement learning (RL) (Biermann & Long, 1996; Levin et al., 2000; Walker et al., 1998a; Singh, Kearns, Litman, & Walker, 1999; Walker, 2000), which have become a standard approach to many AI problems that involve an agent learning to improve performance by interaction with its environment (Sutton &

Barto, 1998; Kaelbling, Littman, & Moore, 1996). More specifically, the MDP and RL formalisms suggest a method for optimizing dialogue policies from sample dialogue data, and have many features well-suited to the problem of dialogue design. These features include the fact that RL is designed to cope gracefully with noisy sensors (such as the ASR), stochastic behavior in the environment (which in this case is the user population), and delayed rewards (which are typical in spoken dialogue systems).[1] The main advantage of this approach is the potential for computing an optimal dialogue policy within a much larger search space, using a relatively small number of training dialogues. The RL approach is more data-efficient because it evaluates actions as a function of state, while the traditional iterative method evaluates entire policies.

Unfortunately, the practical application of RL to the area of spoken dialogue management presents many technical challenges. While the theory of RL is quite advanced, applications have been limited almost exclusively to problems in control, operations research, or game-playing (e.g., (Crites & Barto, 1996; Tesauro, 1995)). Dialogue management represents a rather different type of problem, in which the MDP models a working system's interaction with a population of human users, and RL is used to optimize the system's performance. For this type of application, the amount of training data is severely limited by the requirement that a human interact with the system. Furthermore, the need for exploratory data must be balanced with the need for a functioning system, i.e. each choice that the system has available in a particular context must make sense in that context from the user's perspective.

This paper presents a detailed methodology for using RL to optimize the design of a dialogue management policy based on limited interactions with human users, and experimentally demonstrates the utility of the approach in the context of the NJFun system. At a high level, our RL methodology involves the choice of appropriate performance criteria (i.e., reward measures) and estimates for dialogue state, the deployment of an initial training system that generates deliberately exploratory dialogue data, the construction of an MDP model of user population reactions to different action choices, and the redeployment of the system using the optimal dialogue policy according to this learned or estimated model.

Section 2 describes some of the dialogue policy choices that a dialogue manager must make. Section 3 explains how reinforcement learning can be used to optimize such choices in a fielded dialogue system with human users. Section 4 describes the architecture of the NJFun system, while Section 5 describes how NJFun optimizes its dialogue policy from experimentally obtained dialogue data. Section 6 reports empirical results evaluating the performance of NJFun's learned dialogue policy, and demonstrates that our approach improves NJFun's task completion rate (our chosen measure for performance optimization). Section 6 also presents results establishing the veracity of the learned MDP, and compares the performance of the learned policy to the performance of standard hand-designed policies in the literature. Our results provide empirical evidence that, when properly applied, RL can quantitatively and substantially improve the performance of a spoken dialogue system.

---

1. Other work has explored the use of non-RL learning methods using more immediate kinds of rewards (Chu-Carroll & Brown, 1997; Walker, Rambow, & Rogati, 2001).

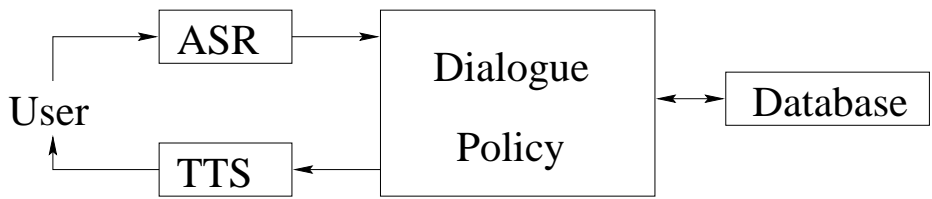

Figure 2: A block diagram representation of a spoken dialogue system. The user gains access to a database by speaking to the system in natural language through the automatic speech recognition system (ASR). The system talks back to the user through a text to speech (TTS) system.

## 2. Dialogue Management in Spoken Dialogue Systems

In a typical spoken dialogue system (shown in block-diagram form in Figure 2), the user speaks to the system in real time through a telephone or microphone, using free-form natural language, in order to retrieve desired information from a back-end such as a database. The user's speech is interpreted through an automatic speech recognizer (ASR), and the system's natural language responses are conveyed to the user via a text-to-speech (TTS) component. The dialogue manager of the system uses a *dialogue policy* to decide what the system should say (or in RL terminology, which *action* it should take) at each point in the dialogue.

For our purposes, an ASR can be viewed as an imperfect, noisy sensor with an adjustable "parameter" (the *language model* or *grammar*) that can be tuned to influence the types of speech recognition mistakes made. In addition to any perceived matches in the utterance, the ASR also returns a score (typically related to log-likelihood under a hidden Markov model) giving a subjective estimate of confidence in the matches found. This score is important in interpreting the ASR results.

Our work concentrates on automating two important types of decisions faced in dialogue policy design, both of which are heavily colored by the ASR facts above. The first type of decisions, of which we have already seen an example, is how much *initiative* the system should allow the user — namely, whether the system at any given point should prompt the user in a relatively open-ended manner (often referred to as *user* initiative) or a relatively restrictive manner (*system* initiative).

The second type of choice we investigate is how conservative the system should be in *confirming* its understanding of the user. After it has applied the ASR to a user utterance, and obtained a value for some attribute of interest (for instance, town = Lambertville), the system must decide whether to confirm the perceived utterance with the user. After the user's response U1 in Figure 1, for example, NJFun must decide whether it should explicitly *confirm* its understanding, as in utterances S2 and S3. NJFun can also simply continue on with the dialogue, as when it does not explicitly confirm that the user wants to find out about wineries. While we might posit that confirmation is unnecessary for high values of the ASR confidence, and necessary for low values, the proper definitions of "high" and "low" would ideally be determined empirically for the current state (for instance, depending on whether there has been difficulty on previous exchanges), and might depend on our measure of system success.

As will be detailed below, in the NJFun system, we identified many different dialogue states for which we wanted to *learn* whether to take user or system initiative for the next prompt. Similarly, we identified many different dialogue states in which we wanted to learn whether to confirm the ASR-perceived user utterance, or not to confirm.[2] We note that there is genuine and spirited debate over choices of initiative and confirmation among dialogue system designers (Walker & Whittaker, 1990; Danieli & Gerbino, 1995; Haller & McRoy, 1998, 1999; Smith, 1998; Walker et al., 1998a). As a simple example, some users enjoy systems that confirm frequently, even if unnecessarily, since it provides confidence that the system is understanding the user. These are not well-understood choices on which there is a prevailing consensus, which is precisely why we wish to automate, in a principled way, the process of making such choices on the basis of empirical data.

## 3. Reinforcement Learning For Dialogue Policy Design

In this section, we describe the abstract methodology we propose to apply RL to dialogue policy design. In the next section, we will describe in detail the instantiation of this methodology in the NJFun system.

In order to apply RL to the design of dialogue policy, it is necessary to define a *state-based* representation for dialogues. By this we simply mean that all or most of the information about the dialogue so far that is relevant for deciding what action the system should take next is contained in a single summarizing entity called the state. One obvious but impractical choice for this state is a transcript or system log of the entire dialogue, which would include the audio so far, the utterances matched by the ASR, the language models used, the confidence scores returned by the ASR, and perhaps many other quantities. In practice, we need to compress this state as much as possible — representing states by the values of a small set of features — without losing information necessary for making good decisions. We view the design of an appropriate state space as *application-dependent*, and a task for a skilled system designer.

Given choices for the state features, the system designer can think in terms of the state space, and appropriate actions to take in each state. We define a *dialogue policy* to be a mapping from the set of *states* in the state space to a set of *actions*. For some states, the proper action to take may be clear (for instance, greeting the user in the start state, or querying the database when all informational attributes are instantiated). For other states, called *choice-states*, there may be multiple reasonable action choices (such as choices of initiative and confirmation). Each mapping from such choice-states to a particular action is a distinct dialogue policy. Typically the system designer uses intuition to choose the best action to take in each choice-state. Our RL-based approach is to instead make these choices by *learning*.

In particular, a dialogue system that explores action choices in a systematic way can learn to optimize its behavior by interacting with representative human users. The system converses with human users to perform a set of representative tasks in the dialogue domain.

---

2. Although not learned in our work, there are obviously many other types of dialogue policy decisions that the system made, e.g., how to present results of database queries (Litman, Pan, & Walker, 1998).

For each dialogue interaction, a scalar performance measure, called a *reward*, is calculated.[3] The resulting dialogue corpus is used to construct a *Markov decision process* (MDP) which models the users' interaction with the system. With this approach, the problem of learning a good dialogue policy is thus reduced to computing the optimal policy for choosing actions in an MDP — that is, the system's goal is to take actions so as to maximize expected reward. The computation of the optimal policy *given* the learned MDP can be done efficiently using standard dynamic programming algorithms (Bertsekas & Tsitsiklis, 1996; Sutton & Barto, 1998).

Since it is difficult to predict next actions, states, and rewards in advance, we build the desired MDP from sample dialogues. Following Singh et al. (1999), we can view a dialogue as a trajectory in the chosen state space determined by the system actions and user responses:

$$s_1 \rightarrow_{a_1, r_1} s_2 \rightarrow_{a_2, r_2} s_3 \rightarrow_{a_3, r_3} \cdots$$

Here $s_i \rightarrow_{a_i, r_i} s_{i+1}$ indicates that at the $i$th exchange, the system was in state $s_i$, executed action $a_i$, received reward $r_i$, and then the state changed to $s_{i+1}$. In our experiments only terminal dialogue states have nonzero rewards. Dialogue sequences obtained from training data can be used to empirically estimate the transition probabilities $P(s'|s, a)$ (denoting the probability of a transition to state $s'$, given that the system was in state $s$ and took action $a$), and the reward function $R(s, a)$ (denoting the expected reward obtained, given that the system was in state $s$ and took action $a$). For example, our estimate of the transition probability is simply the number of times, in all of the dialogues, that the system was in $s$, took $a$, and arrived in $s'$, divided by the number of times the system was in $s$ and took $a$ (regardless of next state). The estimated transition probabilities and reward function constitute an MDP model of the user population's interaction with the system. It (hopefully) captures the stochastic behavior of the users when interacting with the system.

Note that in order to have any confidence in this model, in the sample dialogues the system must have tried many possible actions from many possible states, and preferably many times. In other words, the training data must be *exploratory* with respect to the chosen states and actions. If we never try an allowed action from some state, we cannot expect to know the value of taking that action in that state. Perhaps the most straightforward way of ensuring exploratory training data is to take actions randomly[4]. While this is the approach we will take in NJFun, it requires that we be exceptionally careful in designing the actions allowed at each choice-state, in order to guarantee that the random choices made always result in a dialogue sensible to human users. (Keep in mind that there is no exploration in non choice-states where the appropriate action is already known and fixed by the system designer.) Other approaches to generating exploratory data are possible.

Next, given our MDP, the expected cumulative reward (or *Q-value*) $Q(s, a)$ of taking action $a$ from state $s$ can be calculated in terms of the Q-values of successor states via the

---

3. We discuss various choices for this reward measure later, but in our experiments the reward is always a quantity directly obtainable from the experimental set-up, such as user-satisfaction or task completion.

4. Of course, even with random exploration, it is not possible in practice to explore all states equally often. Some states will occur more often than others. The net effect is that states that occur often will have their actions tried more often than states that occur rarely, and thus the transition probabilities for frequent, and hence potentially important, state-action pairs will be more accurate than the transition probabilities of infrequent state-action pairs.

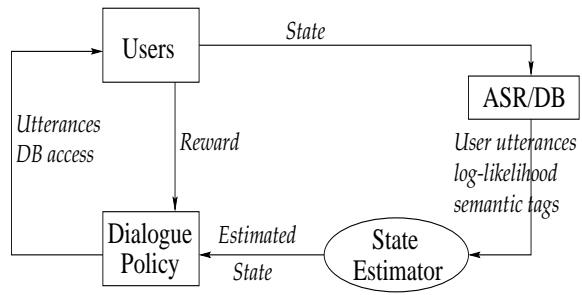

Figure 3: A dialogue system viewed as an MDP. The population of users correspond to the environment whose state is among other things defined by the outputs of the automatic speech recognition (ASR) system and the database (DB). The dialogue policy defines the agent, the state-estimator defines the agent's sensors, and the database actions as well as the possible set of TTS utterances define the agent's action set.

following recursive equation (Watkins, 1989; Sutton & Barto, 1998):

$$Q(s,a) = R(s,a) + \gamma \sum_{s'} P(s'|s,a) \max_{a'} Q(s',a'). \tag{1}$$

where $P(s'|s,a)$ is our estimated transition model and $R(s,a)$ our estimated reward model. Here $0 \leq \gamma \leq 1$ is a discount factor that if set to a value less than one would discount rewards obtained later in time. We found that for NJFun the policy learned was insensitive to reasonable choices of $\gamma$ and therefore we used no discounting, or $\gamma = 1$, for the experiments reported here. The $Q$-values defined by Equation 1 can be estimated to within a desired threshold using the Q-value version of the standard *value iteration* algorithm (Bertsekas & Tsitsiklis, 1996), which iteratively updates the estimate of $Q(s,a)$ based on the current Q-values of neighboring states and stops when the update yields a difference that is below a threshold. Once value iteration is completed, the optimal dialogue policy (according to our estimated model) is obtained by selecting the action with the maximum Q-value at each dialogue state. To the extent that the estimated MDP is an accurate model of the user population, this optimized policy should maximize the reward obtained from *future* users.

While this approach is theoretically appealing, the cost of obtaining sample human dialogues makes it crucial to limit the size of the state space, to minimize data sparsity problems, while retaining enough information in the state to learn an accurate model. If sample data were infinite, the idealized state might include not only the dialogue so far, but also any derived features (e.g. ASR results, log-likelihood scores representing ASR confidence, semantic analysis, the results of database queries, etc.). Yet even a state based on only a small number of features can yield an enormous state space. While others have proposed *simulating* the user interactions to obtain enough training data (Levin et al., 2000; Young, 2000), our approach is to work directly in a small but carefully designed estimated state space (Singh et al., 1999), as shown in Figure 3. By using a minimal state representation to approximate the true state, the amount of data required to learn the optimal dialogue policy for the learned MDP using value iteration can be greatly reduced.

The contribution of this paper is to empirically validate this practical methodology for using reinforcement learning to *build* a dialogue system that optimizes its behavior from human-computer training dialogue data. In a nutshell, our proposed approach is:

1. Choose an appropriate reward measure for dialogues, an appropriate representation for dialogue states, and design a dialogue policy that maps each state to a set of reasonable actions. In many states there may be only one reasonable action.

2. Build an initial state-based *training* system that creates an exploratory data set (one that tries, many times from each choice-state, each of the actions we would like to choose among). Despite being exploratory, this system should still provide the desired basic functionality.

3. Use these training dialogues to build an empirical MDP model on the state space. The transitions of this MDP will be modeling the user population's reactions and rewards for the various system actions.

4. Compute the optimal dialogue policy according to this learned MDP.

5. Reimplement the system using the learned dialogue policy.

The next section details the use of this methodology to design the NJFun system.

## 4. The NJFun System

NJFun is a real-time spoken dialogue system that provides users with information about things to do in New Jersey. NJFun is built using a general purpose platform for spoken dialogue systems (Levin, Pieraccini, Eckert, Fabbrizio, & Narayanan, 1999), with support for modules for automatic speech recognition (ASR), spoken language understanding, text-to-speech (TTS), database access, and dialogue management. NJFun uses the Watson speech recognizer with stochastic language and understanding models trained from example user utterances (Levin et al., 1999; Levin & Pieraccini, 1995), and a TTS system based on concatenative diphone synthesis (Sproat & Olive, 1995). Our mixed-initiative dialogue manager was built using the DMD scripting language (Levin et al., 1999). The NJFun database is populated from the `nj.online` webpage to contain information about the following activity types: amusement parks, aquariums, cruises, historic sites, museums, parks, theaters, wineries, and zoos. NJFun indexes this database using three attributes: activity type, location, and time of day (which can assume values morning, afternoon, or evening).

Informally, the NJFun dialogue manager sequentially queries the user regarding the activity, location and time attributes, respectively. NJFun first asks the user for the current attribute (and possibly the other attributes, depending on the initiative). If the current attribute's value is not obtained, NJFun asks for the attribute (and possibly the later attributes) again. If NJFun still does not obtain a value, NJFun moves on to the next attribute(s). Whenever NJFun successfully obtains a value, it can confirm the value, or move on to the next attribute(s). When NJFun has finished acquiring attributes, it queries the database (using a wildcard for each unobtained attribute value). For any given binding of the three attributes, there may be multiple database matches, which will all be returned to the user. The length of NJFun dialogues ranges from 1 to 12 user utterances before the database query. Although the NJFun dialogues are fairly short (since NJFun asks for an attribute at most twice), the information acquisition part of the dialogue is similar to more complex tasks such as travel planning (Danieli & Gerbino, 1995; Sanderman et al., 1998).[5]

| Grammar | Prompt Type | |
|---|---|---|
| | Open | Directive |
| Restrictive | Doesn't make sense | System Initiative |
| NonRestrictive | User Initiative | Mixed Initiative |

Figure 4: Definition of Initiative for System Prompts.

As discussed above, our methodology for using reinforcement learning to optimize dialogue policy requires that all potential actions for each state be specified. Recall that at some states it is easy for a human to make the correct action choice (e.g., we don't want the system to be able to say "goodbye" in the initial state, as in the simulations of Levin et al. (2000)). We made obvious dialogue policy choices in advance, and used learning only to optimize the difficult choices (Walker et al., 1998a). In NJFun, we restricted the action choices to 1) the type of initiative to use when asking or reasking for an attribute, and 2) whether to confirm an attribute value once obtained. The optimal actions may vary with dialogue state, and are subject to active debate in the literature. The action choices available to NJFun are shown in Figures 5 and 6.

The three types of initiative that the system uses are defined in Figure 4, based on the combination of the wording of the system prompt (*open* versus *directive* (Kamm, 1995))[6], and the type of grammar NJFun uses during ASR (*restrictive* versus *non-restrictive*). The examples in Figure 5 show that NJFun can ask the user about the first two attributes[7] using the three types of initiative. If NJFun uses an open question with a non-restrictive grammar, it is using *user initiative* (e.g., GreetU). The non-restrictive grammar is always used with a user initiative prompt, because the choice of the restrictive grammar does not make sense in that case. If NJFun instead uses a directive prompt with a restricted grammar, the system is using *system initiative* (e.g., GreetS). Here the system calls ASR on the user utterance using a grammar that recognizes only the particular attribute mentioned in the prompt. If NJFun uses a directive question with a non-restrictive grammar, it is using *mixed initiative*, because it allows the user to take the initiative by supplying extra information (e.g., ReAsk1M). The non-restrictive grammar is designed to recognize both the attribute explicitly mentioned in the directive prompt, as well as information offered on the other attributes. The last two rows of the figure show that NJFun always uses system initiative for the third attribute, because at that point the user can only provide the time of day.

---

5. To support continuous use, the system's functionality could be extended in a number of ways such as a larger live database and support for followup questions by the users.

6. While there are other ways of defining initiative (Walker & Whittaker, 1990; Chu-Carroll & Brown, 1997), this operationalization is commonly applied in spoken dialogue systems (Levin et al., 1999).

7. "Greet" is equivalent to asking for the first attribute.

| Action | Prompt | Prompt Type | Grammar |
|--------|--------|-------------|---------|
| GreetS | Welcome to NJFun. Please say an activity name or say 'list activities' for a list of activities I know about. | directive | restrictive |
| GreetU | Welcome to NJFun. How may I help you? | open | nonrestrictive |
| ReAsk1S | I know about amusement parks, aquariums, cruises, historic sites, museums, parks, theaters, wineries and zoos. Please say an activity name from this list. | directive | restrictive |
| ReAsk1M | Please tell me the activity type. You can also tell me the location and time. | directive | nonrestrictive |
| Ask2S | Please say the name of the town or city that you are interested in. | directive | restrictive |
| Ask2U | Please give me more information. | open | nonrestrictive |
| ReAsk2S | Please tell me the name of the town or city that you are interested in. | directive | restrictive |
| ReAsk2M | Please tell me the location that you are interested in. You can also tell me the time. | directive | nonrestrictive |
| Ask3S | What time of the day do you want to go? | directive | restrictive |
| ReAsk3S | Do you want to go in the morning, in the afternoon, or in the evening? | directive | restrictive |

Figure 5: Initiative choices available to NJFun. The first column specifies the names of the actions corresponding to the prompts in the second column. The third column specifies the prompt type and the fourth column specifies the type of grammar used. Actions that can be taken in the same state are grouped together.

NJFun can also vary the actions for confirming each attribute, as shown in Figure 6. If NJFun asks the user to explicitly verify an attribute, it is using *explicit confirmation* (e.g., ExpConf2 for the location, exemplified by S2 in Figure 1). All explicit confirmations are system initiative, as a restrictive yes/no grammar is used, and are generated from templates. For example, the prompt to confirm the time attribute is "Did you say you want to go in the $< time >$?", where $< time >$ is replaced by the perceived value of the time attribute (morning, afternoon, or evening). If NJFun does not generate any confirmation prompt, it is using *no confirmation* (the NoConf action).

Solely for the purposes of controlling its operation (as opposed to the learning, which we discuss in a moment), NJFun internally maintains a representation of the dialogue state, using an *operations vector* of 14 variables. 2 variables track whether the system has greeted the user, and which attribute the system is currently attempting to obtain. For each of the 3 attributes, 4 variables track whether the system has obtained the attribute's value and what the value is, the system's confidence in the value (if obtained), the number of times the system has asked the user about the attribute, and the type of ASR grammar most recently used to ask for the attribute.

| Action | Prompt Template | Prompt Type | Grammar |
|--------|-----------------|-------------|---------|
| ExpConf1 | Did you say you are interested in going to $< activity >$? | directive | restrictive |
| NoConf | - | | |
| ExpConf2 | Did you say you are interested in $< location > $? | directive | restrictive |
| NoConf | - | | |
| ExpConf3 | Did you say you want to go in the $< time >$? | directive | restrictive |
| NoConf | - | | |

Figure 6: Confirmation choices available to NJFun. The first column specifies the names of the actions corresponding to the prompts in the second column. The third column specifies the prompt type and the fourth column specifies the type of grammar used. The prompt for the NoConf (no-confirmation) action is empty.

| Feature | Values | Explanation |
|---------|--------|-------------|
| Greet (G) | 0,1 | Whether the system has greeted the user |
| Attribute (A) | 1,2,3,4 | Which attribute is being worked on |
| Confidence/Confirmed (C) | 0,1,2,3,4 | 0,1,2 for low, medium, and high ASR confidence. 3,4 for explicitly confirmed, and disconfirmed |
| Value (V) | 0,1 | Whether value has been obtained for current attribute |
| Tries (T) | 0,1,2 | How many times current attribute has been asked |
| Grammar (M) | 0,1 | Whether non-restrictive or restrictive grammar was used |
| History (H) | 0,1 | Whether there was trouble on any previous attribute |

Figure 7: State features and values.

The formal state space $\mathcal{S}$ maintained by NJFun for the purposes of learning is much simpler than the operations vector, due to the data sparsity concerns already discussed. The dialogue state space $\mathcal{S}$ contains only 7 variables, as summarized in Figure 7. $\mathcal{S}$ is computed from the operations vector using a hand-designed algorithm. The "Greet" feature tracks whether the system has greeted the user or not (no=0, yes=1). "Attribute" specifies which attribute NJFun is currently attempting to obtain or verify (activity=1, location=2, time=3, done with attributes=4). "Confidence/Confirmed" represents the confidence that NJFun has after obtaining a value for an attribute. The values 0, 1, and 2 represent the lowest, middle and highest ASR confidence values.[8] The values 3 and 4 are set when ASR hears "yes" or "no" after a confirmation question. "Value" tracks whether NJFun has obtained a value for the attribute (no=0, yes=1). "Tries" tracks the number of times that NJFun has asked the user about the attribute. "Grammar" tracks the type of ASR grammar (language model) most recently used to obtain the attribute (0=non-restrictive, 1=restrictive). Finally, "History" represents whether NJFun had trouble understanding the user in the earlier part of the conversation (bad=0, good=1). We omit the full definition, but as an example, when NJFun is working on the second attribute (location), the history variable is set to 0 if NJFun does not have an activity, has an activity but has no confidence in the value, or needed two queries to obtain the activity.

We note that this state representation, in the interests of keeping the state space small, deliberately ignores potentially helpful information about the dialogue so far. For example, there is no state feature explicitly tracking the average ASR score over all user utterances so far, nor do we keep information about the raw feature values for previous states.[9] However, as mentioned above, the goal is to design a small state space that makes enough critical distinctions to support learning. The use of $\mathcal{S}$ reduces the number of states to only 62, and supports the construction of an MDP model that is not sparse with respect to $\mathcal{S}$, even using limited training data.[10] The state space that we utilize here, although minimal, allows us to make initiative decisions based on the success of earlier exchanges, and confirmation decisions based on ASR confidence scores and grammars, as suggested by earlier work (Danieli & Gerbino, 1995; Walker et al., 1998b; Litman, Walker, & Kearns, 1999).

With the state space and action choices precisely defined, we can now detail the *policy class* explored in our experiment, defined to be the set of all deterministic mappings from the states in which the system has a choice to a particular, fixed choice. The state/action mapping representing NJFun's dialogue policy class EIC (Exploratory for Initiative and Confirmation) is shown in Figure 8. For each choice-state, we list the two choices of actions available. (The action choices in boldface are the ones eventually identified as optimal by the learning process, and are discussed in detail later.) Since there are 42 choice-states with 2 action choices each, the total number of unique policies in this class is $2^{42}$. In keeping with the RL methodology described above, our goal is to compute and implement an approximately optimal policy in this large class on the basis of RL applied to exploratory training dialogues.

The policy class in Figure 8 is obtained by allowing a choice of system or user initiative whenever the system needs to ask or reask for an attribute, and by allowing a choice of confirming or simply moving on to the next attribute whenever the system has just obtained a value for an attribute. For example, in the initial state where the user has not yet greeted the user ("Greet" has the value 0), the system has a choice of uttering the system initiative prompt "Please say an activity name or say 'list activities' for a list of activities I know about." or the user initiative prompt "How may I help you?" As another example, choices in confirmation are available at states for which the "Value" feature is 1. In these states, the system can either confirm the attribute value obtained from the ASR, or accept the current binding and move on to the next attribute.

---

8. For each utterance, the ASR output includes not only the recognized string, but also an associated acoustic confidence score. Based on data obtained during system development, we defined a mapping from raw confidence values into 3 approximately equally populated partitions.

9. As discussed above, the system uses its operations vector to store more information, such as the actual values of previous attributes for the eventual database query. As these do not influence future dialogue policy in any way, they are not stored as state features.

10. 62 refers to those states that can actually occur in a dialogue. For example, greet=0 is only possible in the initial dialogue state "0 1 0 0 0 0 0". Thus, all other states beginning with 0 (e.g. "0 1 0 0 1 0 0") will never occur.

| Choice-States | | | | | | | Action Choices |
|---|---|---|---|---|---|---|---|
| G | A | C | V | T | M | H | |
| 0 | 1 | 0 | 0 | 0 | 0 | 0 | GreetS,**GreetU** |
| 1 | 1 | 0 | 0 | 1 | 0 | 0 | **ReAsk1S**,ReAsk1M |
| 1 | 1 | 0 | 1 | 0 | 0 | 0 | NoConf,**ExpConf1** |
| 1 | 1 | 0 | 1 | 0 | 1 | 0 | NoConf,**ExpConf1** |
| 1 | 1 | 1 | 1 | 0 | 0 | 0 | NoConf,**ExpConf1** |
| 1 | 1 | 1 | 1 | 0 | 1 | 0 | NoConf,**ExpConf1** |
| 1 | 1 | 2 | 1 | 0 | 0 | 0 | **NoConf**,ExpConf1 |
| 1 | 1 | 2 | 1 | 0 | 1 | 0 | **NoConf**,ExpConf1 |
| 1 | 1 | 4 | 0 | 0 | 0 | 0 | ReAsk1S,ReAsk1M |
| 1 | 1 | 4 | 0 | 1 | 0 | 0 | **ReAsk1S**,ReAsk1M |
| 1 | 2 | 0 | 0 | 0 | 0 | 0 | Ask2S,**Ask2U** |
| 1 | 2 | 0 | 0 | 0 | 0 | 1 | Ask2S,**Ask2U** |
| 1 | 2 | 0 | 0 | 1 | 0 | 0 | ReAsk2S,**ReAsk2M** |
| 1 | 2 | 0 | 0 | 1 | 0 | 1 | ReAsk2S,**ReAsk2M** |
| 1 | 2 | 0 | 1 | 0 | 0 | 0 | NoConf,**ExpConf2** |
| 1 | 2 | 0 | 1 | 0 | 0 | 1 | NoConf,**ExpConf2** |
| 1 | 2 | 0 | 1 | 0 | 1 | 0 | NoConf,**ExpConf2** |
| 1 | 2 | 0 | 1 | 0 | 1 | 1 | NoConf,ExpConf2 |
| 1 | 2 | 1 | 1 | 0 | 0 | 0 | NoConf,**ExpConf2** |
| 1 | 2 | 1 | 1 | 0 | 0 | 1 | NoConf,**ExpConf2** |
| 1 | 2 | 1 | 1 | 0 | 1 | 0 | **NoConf**,ExpConf2 |
| 1 | 2 | 1 | 1 | 0 | 1 | 1 | NoConf,**ExpConf2** |
| 1 | 2 | 2 | 1 | 0 | 0 | 0 | **NoConf**,ExpConf2 |
| 1 | 2 | 2 | 1 | 0 | 0 | 1 | NoConf,**ExpConf2** |
| 1 | 2 | 2 | 1 | 0 | 1 | 0 | **NoConf**,ExpConf2 |
| 1 | 2 | 2 | 1 | 0 | 1 | 1 | NoConf,ExpConf2 |
| 1 | 2 | 4 | 0 | 0 | 0 | 0 | **ReAsk2S**,ReAsk2M |
| 1 | 2 | 4 | 0 | 0 | 0 | 1 | ReAsk2S,ReAsk2M |
| 1 | 2 | 4 | 0 | 1 | 0 | 0 | ReAsk2S,**ReAsk2M** |
| 1 | 2 | 4 | 0 | 1 | 0 | 1 | ReAsk2S,**ReAsk2M** |
| 1 | 3 | 0 | 1 | 0 | 0 | 0 | NoConf,ExpConf3 |
| 1 | 3 | 0 | 1 | 0 | 0 | 1 | NoConf,ExpConf3 |
| 1 | 3 | 0 | 1 | 0 | 1 | 0 | **NoConf**,ExpConf3 |
| 1 | 3 | 0 | 1 | 0 | 1 | 1 | NoConf,**ExpConf3** |
| 1 | 3 | 1 | 1 | 0 | 0 | 0 | NoConf,**ExpConf3** |
| 1 | 3 | 1 | 1 | 0 | 0 | 1 | NoConf,ExpConf3 |
| 1 | 3 | 1 | 1 | 0 | 1 | 0 | **NoConf**,ExpConf3 |
| 1 | 3 | 1 | 1 | 0 | 1 | 1 | NoConf,**ExpConf3** |
| 1 | 3 | 2 | 1 | 0 | 0 | 0 | NoConf,**ExpConf3** |
| 1 | 3 | 2 | 1 | 0 | 0 | 1 | NoConf,ExpConf3 |
| 1 | 3 | 2 | 1 | 0 | 1 | 0 | **NoConf**,ExpConf3 |
| 1 | 3 | 2 | 1 | 0 | 1 | 1 | NoConf,**ExpConf3** |

Figure 8: EIC Policy Class. Definitions for state features are given in Figure 7.

To execute a particular policy in the policy class EIC, NJFun chooses *randomly* between the two actions for whatever choice-state it is in, thus maximizing exploration and minimizing data sparseness when constructing our MDP model. Note that due to the randomization used for action choice, the prompts in Figures 5 and 6 are designed to ensure the coherence of all possible action sequences.

| State | Action | Turn | Reward |
|-------|--------|------|--------|
| g a c v t m h | | | |
| 0 1 0 0 0 0 0 | GreetU | S1 | 0 |
| 1 1 2 1 0 0 0 | NoConf | - | 0 |
| 1 2 2 1 0 0 1 | ExpConf2 | S2 | 0 |
| 1 3 2 1 0 0 1 | ExpConf3 | S3 | 0 |
| 1 4 0 0 0 0 0 | Tell | S4 | 1 |

Figure 9: Generating the dialogue in Figure 1.

Figure 9 illustrates how the dialogue policy class in Figure 8 generates the dialogue in Figure 1. Each row indicates the state that NJFun is in, the action executed in this state, the corresponding turn in Figure 1, and the reward received. The initial state represents that NJFun will first attempt to obtain attribute 1. NJFun executes GreetU (although as shown in Figure 8, GreetS is also possible), generating the first utterance in Figure 1. After the user's response, the next state represents that NJFun has now greeted the user and obtained the activity value with high confidence, by using a non-restrictive grammar. NJFun then chooses the NoConf action, so it does not attempt to confirm the activity, which causes the state to change but no prompt to be generated. The third state represents that NJFun is now working on the second attribute (location), that it already has this value with high confidence (location was obtained with activity after the user's first utterance), and that the dialogue history is good.[11] This time NJFun chooses the ExpConf2 action, and confirms the attribute with the second NJFun utterance, and the state changes again. The processing of time is similar to that of location, which leads NJFun to the final state, where it performs the action "Tell" (corresponding to querying the database, presenting the results to the user, and asking the user to provide a reward). Note that in NJFun, the reward is always 0 except at the terminal state, as shown in the last column of Figure 9.

Figure 10 illustrates another NJFun dialogue that was generated using the EIC policy class. Note that the dialogues in Figures 1 and 10 instantiate two *different* dialogue policies in the EIC policy class. For example, although NJFun begins both dialogues in the first state in Figure 8, NJFun executes GreetU in one dialogue and GreetS in the other.

## 5. Experimentally Optimizing a Policy

We collected experimental dialogues for both training and testing our system. To obtain training dialogues, we implemented NJFun using the EIC dialogue policy class described

---

11. Recall that only the current attribute's features are in the state. However, the operations vector contains information regarding previous attributes.

S1: Welcome to NJFun. Please say an activity name or say 'list activities' for a list of activities I know about.

U1: I would like to visit a historic site in Stanhope this morning.
(*ASR output: I would like to the zoo historic sites in stanhope historic*)

S2: Did you say you are interested in going to a zoo?

U2: No.

S3: I know about amusement parks, aquariums, cruises, historic sites, museums, parks, theaters, wineries, and zoos. Please say an activity name from this list.

U3: I would like to visit a historic site. (*ASR output: I would like to visit historic sites*)

S4: Did you say you are interested in going to a historic site?

U4: Yes.

S5: Please give me more information.

U5: *(user says nothing)*

S6: Please tell me the location that you are interested in. You can also tell me the time.

U6: Visit a historic. (*ASR output: What historic*)

S7: What time of the day do you want to go?

U7: Stanhope. (*no ASR output*)

S8: Do you want to go in the morning, in the afternoon, or in the evening?

U8: Morning.

S9: Did you say you want to go in the morning?

U9: Yes.

S10: I found 11 historic sites in New Jersey that are open in the morning. The first 3 [...] Would you like to hear more?

U10: No.

S11: Thank you for using the system. Please give me feedback by saying 'good', 'so-so', or 'bad'.

U11: Bad.

Figure 10: Another example dialogue with NJFun.

in Section 4. We used these dialogues to build an empirical MDP, and then computed the optimal dialogue policy in this MDP (as described in Section 3). In this section we describe our experimental design and the learned dialogue policy. In the next section we present results from testing our learned policy and show that it improves task completion rates, the performance measure we chose to optimize.

Experimental subjects were AT&T employees not associated with the NJFun project. There were 54 subjects for training and 21 for testing. Subjects were distributed so the training and testing pools were balanced for gender, English as a first language, and expertise with spoken dialogue systems.[12] Training subjects were informed at the beginning of the experiment that NJFun might change its behavior during the experiment, via a set of web-based instructions (see Appendix A).

---

12. Subsequent analyses indicated that system performance did not depend significantly on any of these factors.

- Task 1. You are bored at home in Morristown on a rainy afternoon. Use NJFun to find a museum to go to.

- Task 2. You live in Cape May and want to take some friends on an evening cruise. Use NJFun to find out what your options are.

- Task 3. You have lived in Stanhope for many years but have never managed to visit its historic sites. Today you are feeling virtuous. Use NJFun to find out what you can see this morning.

- Task 4. You feel thirsty and want to do some winetasting in the morning. Are there any wineries close by your house in Lambertville?

- Task 5. After a hard day of work at AT&T in Florham Park, you would like to relax with an evening at the theatre. Use NJFun to find out if it is possible to see a show near Florham Park.

- Task 6. You live in Jersey City, and want to spend the afternoon enjoying nature as the weather is beautiful. Are there any parks nearby?

Figure 11: Task scenarios.

During both training and testing, subjects carried out free-form conversations with NJFun to complete the six application tasks in Figure 11. For example, the task executed by the user in Figure 1 was Task 4 in Figure 11. Subjects read each task description by going to a separate web page for each task (accessible from the main experimental web page), then called NJFun from their office phone. At the end of the task, NJFun asked for feedback on their experience (e.g., utterance S4 in Figure 1). Users then hung up the phone and filled out a user survey on the web, shown in Figure 12. Possible responses for questions 1 and 2 are shown. The answers to the first question (*good, so-so, bad*) are mapped to 1, 0, and -1, respectively. For the remaining questions, users indicated the strength of their agreement on a 5 point Likert scale (Jack, Foster, & Stentiford, 1992), with the responses (*strongly agree, somewhat agree, neither agree nor disagree, somewhat disagree, strongly disagree*), which are mapped to 5 through 1, respectively.

As dictated by Step 2 of the RL methodology described in Section 3, we first built a *training* version of the system, using the EIC state space and action choices outlined in the preceding section, that used *random exploration*. By this we mean that in any state for which we had specified a choice of system actions, the training system chose randomly among the allowed actions with uniform probability. We again emphasize the fact that the allowed choices were designed in a way that ensured that any dialogue generated by this exploratory training system was intuitively sensible to a human user, and permitted the successful completion of any task the system was intended to perform. Nevertheless, it is important to note that over their multiple calls to the system, training users may have effectively experienced multiple dialogue policies (as induced by the random exploration), while test users experienced a single, fixed, deterministic policy.

- Please repeat (or give) your feedback on this conversation. (*good, so-so, bad* )

- Did you complete the task and get the information you needed? (*yes, no* )

- In this conversation, it was easy to find the place that I wanted.

- In this conversation, I knew what I could say at each point in the dialogue.

- In this conversation, NJFun understood what I said.

- Based on my current experience with using NJFun, I'd use NJFun regularly to find a place to go when I'm away from my computer.

Figure 12: User survey.

The training phase of the experiment resulted in 311 complete dialogues (not all subjects completed all tasks), for which NJFun logged the sequence of states and the corresponding executed actions. The shortest and longest dialogues obtained had 3 and 11 user utterances, respectively. In our training set, the number of samples per state for the initial ask choices are:

```
0 1 0 0 0 0 0   GreetS=155   GreetU=156
1 2 0 0 0 0 0   Ask2S=93     Ask2U=72
1 2 0 0 0 0 1   Ask2S=36     Ask2U=48
```

Such data illustrates that the random action choice method of exploration led to a fairly balanced action distribution per state. Similarly, the small state space, and the fact that we only allowed 2 action choices per state, prevented a data sparseness problem. This is important because the optimal dialogue policy obtained via RL is unreliable at infrequently visited states. The first state in Figure 8, the initial state for every dialogue, was the most frequently visited state (with 311 visits). Only 8 states that occur near the end of a dialogue were visited less than 10 times.

The logged data was then used to construct the empirical MDP. As we have mentioned, the measure we chose to optimize is a binary reward function based on the strongest possible measure of task completion, called **Binary Completion**, that takes on value 1 if NJFun queries the database using exactly the attributes specified in the task description, and -1 otherwise. Since system logs could be matched with which of the six tasks the user was attempting, it was possible to directly compute from the system logs whether or not the user had completed the task. By "completed" we mean binding all three attributes (activity type, location, and time of day) to the exact values specified in the task description given on the associated web page. In this way, each training dialogue was automatically labeled by a +1 in the case of a completed task, or −1 otherwise. We note that this definition of task completion guarantees that the user heard all and only the database entries matching the task specifications. Relaxations of this reward measure, as well as other types of measures that could have been used as our reward measure, are discussed in the next section.

Finally, we computed the optimal dialogue policy in this learned MDP using Q-value iteration (cf. Section 3). The action choices constituting the learned policy are in boldface in Figure 8. Note that no choice was fixed for several states (e.g., "1 1 4 0 0 0 0"), meaning that the Q-values were identical after value iteration. Thus, even when using the learned policy, NJFun still sometimes chooses randomly between certain action pairs.

Intuitively, the learned policy says that the optimal use of initiative is to begin with user initiative, then back off to either mixed or system initiative when reasking for an attribute. Note, however, that the specific backoff method differs with attribute (e.g., system initiative for attribute 1, but generally mixed initiative for attribute 2). With respect to confirmation, the optimal policy is to mainly confirm at lower confidence values. Again, however, the point where confirmation becomes unnecessary differs across attributes (e.g., confidence level 2 for attribute 1, but sometimes lower levels for attributes 2 and 3), and also depends on other features of the state besides confidence (e.g., grammar and history). This use of ASR confidence by the dialogue policy is more sophisticated than previous approaches, e.g. (Niimi & Kobayashi., 1996; Litman & Pan, 2000). NJFun can learn such fine-grained distinctions because the optimal policy is based on a comparison of $2^{42}$ possible exploratory policies. Both the initiative and confirmation results suggest that the beginning of the dialogue was the most problematic for NJFun. Figure 1 is an example dialogue using the optimal policy.

## 6. Experimentally Evaluating the Optimized Policy

For the testing phase, NJFun was reimplemented to use the (now deterministic) learned policy. 21 test subjects then performed the same six tasks used during training, resulting in 124 complete test dialogues. The primary empirical test of the proposed methodology is, of course, the extent and statistical significance of the improvement in the allegedly optimized measure (binary task completion) from the training to test populations. In fact, task completion as measured by **Binary Completion** does increase, from 52% in training to 64% in testing. The following sections are devoted to the analysis of this test, as well as several related tests.

### 6.1 Comparing the Learned Policy to the Training Policy

Table 1 summarizes the training versus testing performance of NJFun, for various evaluation measures. Recall that in the 311 training dialogues, NJFun used randomly chosen policies in the EIC policy class. In the 124 testing dialogues, NJFun used the single learned policy. Although the learned policy was optimized for only the task success measure **Binary Completion**, many types of measures have been used to evaluate dialogue systems (e.g., task success, dialogue quality, efficiency, usability (Danieli & Gerbino, 1995; Kamm et al., 1998)). We thus evaluate the performance of the learned policy with respect to both the original reward measure and a number of other potential reward measures that we did not optimize the test system for.

Perhaps our most important results are summarized in the first two rows of Table 1. In the first row, we summarize performance for the **Binary Completion** reward measure, discussed in the preceding section. The average value of this reward measured across the 311 dialogues generated using the randomized training system was 0.048 (recall the range is −1 to 1), while the average value of this same measure across the 124 dialogues using the learned test system was 0.274, an improvement that has a p-value of 0.059 in a standard two-sample t-test over subject means.[13] This result corresponds to an improvement from a 52% completion rate among the training dialogues to a 64% completion rate among the testing dialogues.

| Evaluation Measure | Train | Test | $\Delta$ | p-value |
|---|---|---|---|---|
| Binary Completion | 0.048 | 0.274 | 0.226 | 0.059 |
| Weak Completion | 1.72 | 2.18 | 0.46 | 0.029 |
| ASR | 2.48 | 2.67 | 0.19 | 0.038 |
| Web feedback | 0.18 | 0.11 | $-0.07$ | 0.42 |
| Easy | 3.38 | 3.39 | 0.01 | 0.98 |
| What to say | 3.71 | 3.64 | $-0.07$ | 0.71 |
| NJFun understood | 3.42 | 3.52 | 0.1 | 0.58 |
| Reuse | 2.87 | 2.72 | $-0.15$ | 0.55 |

Table 1: Train versus test performance for various evaluation measures. The first column presents the different measures considered (see text for detail); the second column is the average value of the measure obtained in the training data; the third column is the average value obtained in the test data; the fourth column shows the difference between the test average and the train average (a positive number is a "win", while a negative number is a "loss"); the fifth column presents the statistical significance value obtained using the standard t-test.

The second row of Table 1 shows that performance also improves from training to test for the closely related measure **Weak Completion**[14]. **Weak Completion** is a relaxed version of task completion that gives partial credit: if all attribute values are either correct or wildcards, the value is the sum of the correct number of attributes. Otherwise, at least one attribute is wrong (e.g., the user says "Lambertville" but the system hears "Morristown"), and the value is -1. The motivation for this more refined measure is that reward -1 indicates that the information desired was not contained in the database entries presented to the user, while non-negative reward means that the information desired was present, but perhaps buried in a larger set of irrelevant items for smaller values of the reward. The training dialogue average of weak completion was 1.72 (where the range is $-1$ to 3), while the test dialogue average was 2.18. Thus we have a large improvement, this time significant at the 0.029 level. We note that the policy dictated by optimizing the training MDP for binary completion (which was implemented in the test system), and the policy dictated by optimizing the training MDP for weak completion (which was not implemented) were very similar, with only very minor differences in action choices.

---

13. Conventionally, a p-value of less than .05 is considered to be statistically significant, while p-values less than .10 are considered indicative of a statistical trend.

14. We emphasize that this is the *improvement* in *weak* completion in the system that was designed to optimize *binary* completion — that is, we only fielded a single test system, but examined performance changes for several different evaluation measures which could also have been used as our reward measure.

The measure in the third row, **ASR**, is another variation of **Binary Completion**. However, instead of evaluating task success, **ASR** evaluates dialogue quality. In particular, **ASR** approximates speech recognition accuracy for the database query, and is computed by adding 1 for each correct attribute value and .5 for every wildcard. Thus, if the task is to go winetasting near Lambertville in the morning, and the system queries the database for an activity in New Jersey in the morning, **Binary Completion**=-1, **Weak Completion**=1, and **ASR**=2. Table 1 shows that the average value of ASR increased from 2.48 during training to 2.67 during testing (where the range is 0 to 3), a significant improvement ($p <$ 0.04). Again, this improvement occurred even though the learned policy used for testing was not optimized for **ASR**.

The three measures considered so far are *objective* reward measures, in the sense that the reward is precisely defined as a function of the system log on a dialogue, and can be computed directly from this log. We now examine how performance changes from training to test when a set of *subjective* usability measures (provided by the human user following each dialogue) are considered. Recall that each dialogue task was accompanied by the web survey in Figure 12. The measure **Web feedback** is obtained from the first question in this survey (recall the range is −1 to 1). The measures **Easy**, **What to say**, **NJFun understood** and **Reuse** are obtained from the last four questions (recall the range is 1 to 5). Since we did not optimize for any of these subjective measures, we had no *a priori* expectations for improvement or degradation. The last five rows of Table 1 shows we in fact did not find any statistically significant changes in the mean in either direction for these measures. However, we observed a curious *move to the middle* effect in that a smaller fraction of users had extremely positive or extremely negative things to say about our test system than did about the training system. Figure 13, which shows the entire distribution of the values for both the train and test systems for these subjective measures, shows that in optimizing the test system for the task completion measure, we seem to have consistently shifted weight away from the tails of the subjective measures, and towards the intermediate values. Although we have no firm explanation for this phenomenon, its consistency (it occurs to varying degree for all 5 subjective measures) is noteworthy.

In sum, our empirical results have demonstrated improvement in the optimized task completion measure, and also improvement in two non-optimized (but related) objective measures. In contrast, our results show no statistically significant changes for a number of non-optimized subjective measures, but an interesting move to the middle effect.

## 6.2 Effect of Expertise

In addition to the task-independent performance changes from training to testing policy just discussed, there were also task-dependent performance changes. For example, there was a significant interaction effect between policy and task (p<.01) when performance was evaluated for **Binary Completion**.[15] We believe that this could be the effect of user expertise with the system since previous work suggests that novice users perform comparably to experts after only two tasks (Kamm et al., 1998). Since our learned policy

---

15. Our experimental design consisted of two factors: the within-group factor *policy* and the between-groups factor *task*. We use a two-way analysis of variance (ANOVA) to compute interaction effects between policy and task.

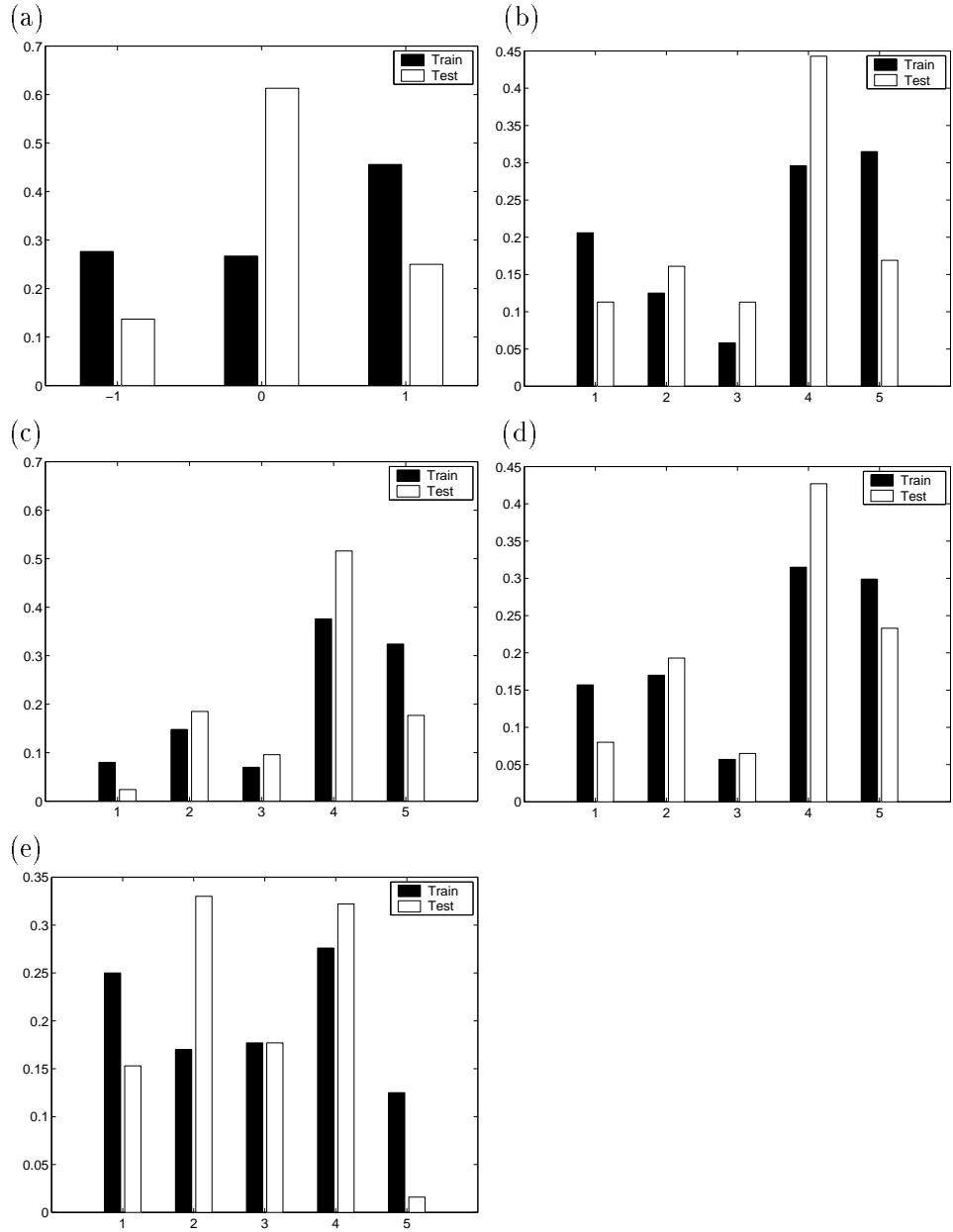

Figure 13: Distributions of the subjective measures. (a) Web feedback. (b) Easy. (c) What to say. (d) NJFun understood. (e) Reuse.

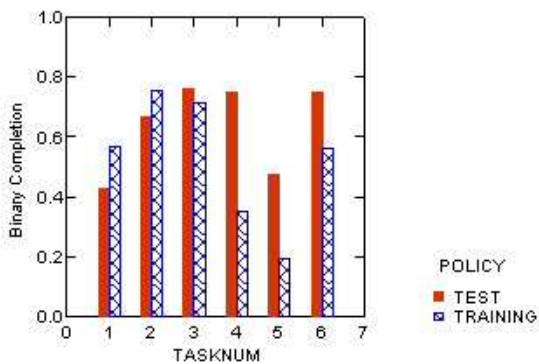

Figure 14: Interaction effects between task and policy. The bar charts show the binary
completion rate for the six tasks (in the order they were presented) for the test
and train policies. The test policy performance is better for the last four tasks
while the train policy performance is better on the first two tasks, providing
evidence that the learned test policy is slightly optimized for expert users.

was based on six tasks with each user, it is possible that the learned policy is slightly
optimized for expert users. To explore this hypothesis, we divided our corpus into dialogues
with "novice" (tasks 1 and 2) and "expert" (tasks 3-6) users. We found that the learned
policy did in fact lead to a large and significant improvement in **Binary Completion** for
experts, increasing the number of completed dialogues from 46% during training to 69%
during testing ($p<.001$). In contrast, there was a non-significant degradation for novices
(train=66%, test=55%, $p<.3$). In particular, as shown in Figure 14, the test means are
lower than the train means for the first two tasks, but higher for the last four tasks. This
is appropriate for a system that has repeat usage; however should it be the case that our
system is primarily used by novice users, the system might need to be retrained.

## 6.3 Comparison to Hand Designed Policies

Although the results presented so far indicate an improvement from training to testing, a
potential limitation is that using a set of policies in the EIC class may not be the best
baseline for comparison to our learned policy. A more standard alternative would be com-
parison to the very best hand-designed fixed policy. However, there is no agreement in the
literature, nor amongst the authors, as to what the best hand-designed policy might have
been. Nevertheless, it is natural to ask how our optimized system compares to systems
employing a dialogue policy picked by a human expert. Although implementing a number
of hand-picked policies, gathering dialogues from them, and comparing to our learned sys-
tem would be time-consuming and expensive (and in fact, is exactly the kind of repeated,
sequential implement-and-test methodology we are attempting to replace), our training sys-
tem provides a convenient and mathematically sound proxy. In this section we show that the performance of the learned policy is better than several "standard" fixed policies, by computing the reward for all the trajectories in the empirical MDP that are consistent with each alternative policy. Then, because each of these alternatives has only a handful of consistent trajectories in the MDP, in the next section we present an analysis of the MDP's accuracy.

Since our training dialogues are generated making *random* choices, any dialogue in the training set that is *consistent* with a policy $\pi$ in our policy class provides an unbiased Monte Carlo trial of $\pi$. By consistent we mean that all the random choices in the dialogue agree with those dictated by $\pi$. We can average the rewards over the consistent training dialogues to obtain an unbiased estimate of the return of $\pi$.

| Policy | # of Trajs. | Emp. Avg. | MDP Value | p-value |
|---|---|---|---|---|
| Test | 12 | 0.67 | 0.534 | |
| SysNoconfirm | 11 | $-0.08$ | 0.085 | 0.06 |
| SysConfirm | 5 | $-0.6$ | 0.006 | 0.01 |
| UserNoconfirm | 15 | $-0.2$ | 0.064 | 0.01 |
| UserConfirm | 11 | 0.2727 | 0.32 | 0.30 |
| Mixed | 13 | $-0.077$ | 0.063 | 0.06 |

Table 2: Comparison to standard policies. Here we compare our test policy with several standard policies using the Monte Carlo method. The first column presents the different policies considered (see text for detail); the second column shows the number of consistent trajectories in the training data; the third column shows the empirical average reward on these consistent trajectories; the fourth column shows the estimated value of the policy according to our learned MDP, and the fifth column shows the statistical significance (p-value) of the policy's loss with respect to the test policy.

Table 2 compares the performance of our learned test system, on the **Binary Completion** reward measure, to 5 fixed policies in our class that are common choices in the dialogue systems literature, or that were suggested to us by dialogue system designers. The SysNoconfirm policy always uses system initiative and never confirms; the SysConfirm policy always uses system initiative and confirms; the UserNoconfirm policy always uses user initiative and never confirms; the UserConfirm policy always uses user initiative and confirms; the Mixed policy varies the initiative during the dialogue. For all but the User-Confirm policy, the test policy is better with a significance near or below the 0.05 level, and the difference with UserConfirm is not significant. (Not surprisingly, the fixed UserConfirm policy that fared best in this comparison is most similar to the policy we learned.) Thus, in addition to optimizing over a large class of policy choices that is considerably more refined than is typical, the reinforcement learning approach outperforms a number of natural standard policies.

## 6.4 The Goodness of Our MDP

Finally, we can ask whether our estimate of state was actually a good estimate or whether we have simply been fortunate — that is, whether our MDP might have actually been a rather poor predictor of the value of actions, but that we happened to have nevertheless chosen a good policy by chance. As some closing evidence against this view, we offer the results of a simple experiment in which we randomly generated many (deterministic) policies in our policy class. For each such policy $\pi$, we used the training dialogues consistent with $\pi$ to compute an unbiased Monte Carlo estimate $\hat{R}_\pi$ of the expected (binary completion) return of $\pi$ (exactly as was done for the hand-picked "expert" policies in Table 2). This estimate was then paired with the value $R_\pi$ of $\pi$ (for the start state) in the learned MDP. If the MDP were a perfect model of the user population's responses to system actions, then the Monte Carlo estimate $\hat{R}_\pi$ would simply be a (noisy) estimate of $R_\pi$, the correlation between these two quantities would be significant (but of course dependent on the number of samples in the Monte Carlo estimate), and the best-fit linear relationship would be simply $\hat{R}_\pi = R_\pi + Z$ (slope 1 and intercept 0), where $Z$ is a normally distributed noise variable with adjustable mean and variance decreasing as the number of consistent trajectories increases. At the other extreme, if our MDP had no relation to the user population's responses to system actions, then $\hat{R}_\pi$ and $R_\pi$ would be uncorrelated, and the best we could do in terms of a linear fit would be $\hat{R}_\pi = Z$ (slope and intercept 0) — that is, we ignore $R_\pi$ and simply model $\hat{R}_\pi$ as noise. The results summarized in Table 3 indicate that we are much closer to the former case than the latter. Over the 1000 random policies $\pi$ that we generated, the correlation between $\hat{R}_\pi$ and $R_\pi$ was positive and rejected the null hypothesis that the variables are uncorrelated well below the 0.01 level of significance; furthermore, the least squares linear fit gave a slope coefficient close to 1.0 and a y-intercept close to 0, as predicted by the idealized case above.

## 7. Discussion

In this paper we presented a practical methodology for applying reinforcement learning to the problem of optimizing dialogue policy design in spoken dialogue systems. Our methodology takes a relatively small number of exploratory dialogues, and directly computes the apparent optimal policy within a space of perhaps thousands of policies, instead of performing a sequence of implementations of only a handful of particular policies. We have used this method to construct a training version of the NJFun spoken dialogue system, and have empirically demonstrated improved performance in NJFun after optimization. In a controlled experiment with human users using NJFun, we verified significant improvements in the reward measure for which the optimization was performed. We also showed that there were significant improvements for several other objective reward measures (even though the test policy was not optimized for these measures), but no improvements for a set of subjective measures (despite an interesting change in their distributions). Finally, we showed that the learned policy is not only better than the non-deterministic EIC policy class, but also better than other fixed choices proposed in the literature. Our results demonstrate that the application of reinforcement learning allows one to empirically optimize a system's dialogue policy by searching through a much larger search space than can be explored with more traditional methods.

| # of Trajs. | # of Policies | Corr. Coeff. | p-value | Slope | Inter. |
|---|---|---|---|---|---|
| > 0 | 1000 | 0.31 | 0.00 | 0.953 | 0.067 |
| > 5 | 868 | 0.39 | 0.00 | 1.058 | 0.087 |
| > 10 | 369 | 0.5 | 0.00 | 1.11 | 0.11 |

Table 3: A test of MDP accuracy. We generated 1000 deterministic policies randomly. For each policy we computed a pair of numbers: its estimated value according to the MDP, and its value based on the trajectories consistent with it in the training data. The number of consistent trajectories varied with policy. The first row is for all 1000 policies, the second row for all policies that had at least 5 consistent trajectories, and the last row for all policies that had at least 10 consistent trajectories. The reliability of the empirical estimate of a policy increases with increasing number of consistent trajectories. The third column presents the correlation coefficient between the empirical and MDP values. The fourth column presents the statistical significance of the correlation coefficient. The main result is that the hypothesis that these two sets of values are uncorrelated can be soundly rejected. Finally, the last two columns present the slope and intercept resulting from the best linear fit between the two sets of values.

Reinforcement learning has been applied to dialogue systems in previous work, but our approach differs from previous work in several respects. Biermann and Long (1996) did not test reinforcement learning in an implemented system, and the experiments of Levin et al. (2000) utilized a simulated user model. Walker et al. (1998a)'s methodology is similar to that used here, in testing reinforcement learning with an implemented system with human users. Walker et al. (1998a) explore initiative policies and policies for information presentation in a spoken dialogue system for accessing email over the phone. However that work only explored policy choices at 13 states in the dialogue, which conceivably could have been explored with more traditional methods (as compared to the 42 choice states explored here).

We also note that our learned policy made dialogue decisions based on ASR confidence in conjunction with other features, and also varied initiative and confirmation decisions at a finer grain than previous work; as such, our learned policy is not a standard policy investigated in the dialogue system literature. For example, we would not have predicted the complex and interesting back-off policy with respect to initiative when reasking for an attribute.

Our system and experiments have begun to address some of the challenges spoken dialogue systems present to the prevailing theory and application of RL (e.g., balancing the competing concerns of random exploration with user experience in a fielded training system; keeping the state space as small as possible in order to make learning data-efficient, while retaining all information necessary for decision-making). However, other challenges remain to be addressed. We do not provide a general methodology for reducing the state space to a manageable size. Furthermore, in our work our learned MDP model is at best an approximation, we may be introducing the problem of hidden state or partial observability into the problem of choosing optimal actions in each state. For situations with hidden state a richer POMDP model is often more appropriate (Kaelbling et al., 1996). Roy, Pineau, and Thrun (2000) are currently exploring whether a POMDP-style approach can yield MDP-like speeds in a spoken dialogue system for a robot, where state is used to represent the user's intentions rather than the system's state.

As future work, we wish to understand the aforementioned results on the subjective measures, explore the potential difference between optimizing for expert users and novices, automate the choice of state space and reward for dialogue systems (which in our methodology is assumed to be given), investigate the use of a learned reward function (Walker et al., 1998a), and explore the use of more informative non-terminal rewards.

## Acknowledgments

The authors thank Fan Jiang for his substantial effort in implementing NJFun, Wieland Eckert, Esther Levin, Roberto Pieraccini, and Mazin Rahim for their technical help, Julia Hirschberg for her comments on a draft of this paper, and David McAllester, Richard Sutton, Esther Levin and Roberto Pieraccini for helpful conversations.

## Appendix A. Experimental Instructions

### NJFun (The New Jersey Place-to-go Recommender)

### General Description

NJFun is an experimental spoken dialogue system that allows you to access a database of things to do in New Jersey via a telephone conversation. You will be asked to call NJFun to do 6 different tasks. You should try to do each task as efficiently as you can. Note that you will be speaking to a different version of NJFun during each phone call, and that NJFun might even vary its behavior within a single phone call.

Instructions for calling NJFun can be found at each task scenario. Please read through the instructions before calling. On rare occasions, you may get an apparently dead line when you call. This indicates that all lines are busy. If this occurs, hang up and call later. Also, PLEASE DO NOT USE A SPEAKER PHONE.

At the end of each task, you will be asked to say "good", "so-so", or "bad" , in order to provide feedback on your phone call with NJFun. PLEASE DO NOT HANG UP THE PHONE BEFORE PROVIDING THIS FEEDBACK. After you hang up the phone, there will also be a few brief questions for you to answer. Even if NJFun aborted before you could complete the task, PLEASE FINISH THE SURVEY and continue to the next task. Once you have finished ALL of the tasks, there will also be an opportunity for you to provide further comments.

If you have any problems during the experiment, call Diane at 973-360-8314, or Satinder at 973-360-7154.

Thank you for participating in this experiment!

## Task Scenarios

You have 6 tasks to try in this experiment. You should do one task at a time, in the prescribed order. After you finish each task and have provided your feedback, hang up the phone and finish the survey for that task. Once you have finished ALL of the tasks, please provide any final comments.

- Click here to try Task 1

- Click here to try Task 2

- Click here to try Task 3

- Click here to try Task 4

- Click here to try Task 5

- Click here to try Task 6

- Click here to provide final comments